# Coordinated Semantic Alignment and Evidence Constraints for Retrieval-Augmented Generation with Large Language Models

Xin Chen
Cornell University
Ithaca, USA

Saili Uday Gadgil
Virginia polytechnic institute and state University
Blacksburg, USA

Jiarong Qiu*
University of Southern California
Los Angeles, USA

*Abstract*-Retrieval augmented generation mitigates limitations of large language models in factual consistency and knowledge updating by introducing external knowledge. However, practical applications still suffer from semantic misalignment between retrieved results and generation objectives, as well as insufficient evidence utilization. To address these challenges, this paper proposes a retrieval augmented generation method that integrates semantic alignment with evidence constraints through coordinated modeling of retrieval and generation stages. The method first represents the relevance between queries and candidate evidence within a unified semantic space. This ensures that retrieved results remain semantically consistent with generation goals and reduces interference from noisy evidence and semantic drift. On this basis, an explicit evidence constraint mechanism is introduced. Retrieved evidence is transformed from an implicit context into a core control factor in generation. This restricts the expression scope of generated content and strengthens dependence on evidence. By jointly modeling semantic consistency and evidence constraints within a unified framework, the proposed approach improves factual reliability and verifiability while preserving natural language fluency. Comparative results show stable improvements across multiple generation quality metrics. This confirms the effectiveness and necessity of coordinated semantic alignment and evidence constraint modeling in retrieval augmented generation tasks.

## I. INTRODUCTION

Against the backdrop of rapid advances in large language models, generative approaches have demonstrated strong capabilities in language modeling and knowledge transfer across tasks such as question answering, summarization, dialogue, and decision support[1]. However, when applied to scenarios that require high domain expertise, strict factual grounding, or strong timeliness, these models often reveal critical limitations. These include unclear knowledge boundaries, insufficient fact traceability, and a disconnect between generated content and real evidence. In complex information environments, models tend to rely on internal parameters to produce outputs that appear plausible but lack factual support. This issue restricts the applicability of large language models in high-risk settings and raises higher demands on their reliability and controllability.

Retrieval augmented generation introduces external knowledge retrieval into the generation process to provide models with updatable and traceable sources of factual information. This paradigm is widely regarded as an effective way to mitigate the above issues[2]. By combining language generation with information retrieval, models can reference real corpora during response generation, which helps reduce hallucinations and improve factual consistency. However, existing retrieval augmented methods still face major challenges in practice. On the one hand, the semantic alignment between retrieved results and generation objectives is often unstable. Evidence with low relevance or high noise can be introduced and interfere with model decisions. On the other hand, retrieved evidence is usually treated as implicit context without explicit constraints. As a result, models may still deviate from the evidence during generation, leading to insufficient or selective use of retrieved information[3].

From a semantic perspective, the weak alignment between retrieval and generation fundamentally arises from the mismatch between their modeling objectives. Retrieval focuses on surface-level similarity or keyword matching, while generation relies on deep semantic representations and contextual reasoning. This discrepancy makes it difficult for retrieved texts to precisely match generation needs in structure, perspective, or granularity[4]. Without effective semantic alignment mechanisms, models may fail to absorb key information from relevant evidence, even when such evidence is formally related. This directly affects the quality of generated outputs. Therefore, establishing stable and consistent semantic mappings between the retrieval and generation stages is a critical issue that retrieval-augmented generation methods must address.

At the same time, semantic alignment alone is insufficient to guarantee the reliability of generated results. In the absence of explicit constraints, models may freely combine multiple pieces of evidence or perform implicit inference. This can introduce information that is not directly supported by evidence. Such evidence drift is particularly prominent in complex queries and multi-hop reasoning scenarios. It undermines the advantages of retrieval augmented generation in terms of interpretability and controllability. Introducing explicit evidence constraints to restrict content selection, information fusion, and expression scope is therefore essential.

This enhances the verifiability and consistency of generated outputs. It also strengthens evidence dependence and supports downstream auditing and accountability[5].

Under this context, retrieval augmented generation methods that integrate semantic alignment with evidence constraints have substantial research value and practical significance. Jointly modeling semantic consistency and evidence constraints within a unified framework can effectively alleviate the structural disconnect between retrieval and generation. External knowledge is thus transformed from an optional reference into a core driving factor. This approach promotes a shift from heuristic integration toward systematic modeling in retrieval augmented generation[6]. It lays a foundation for building large language models that are more reliable, controllable, and aligned with real-world application demands. Moreover, this direction provides a new modeling paradigm for complex knowledge-intensive tasks and contributes to the long-term reliability and societal value of intelligent systems.

## II. Methodological Foundations

The framework proposed in this study is built upon a set of complementary methodological foundations spanning memory-augmented architectures, explainable representation learning, modular and structural adaptation, as well as adaptive control for reliable retrieval augmented generation. At its core, explicit memory-driven mechanisms, such as auxiliary rationale memory—enable the systematic storage, retrieval, and use of intermediate rationales throughout the generation process. This foundation is essential for maintaining factual consistency, transparent reasoning, and verifiable evidence integration in retrieval-augmented systems [7].

Explainable representation learning and neural attention mechanisms provide another critical pillar. By supporting fine-grained alignment between model input, retrieved knowledge, and output, these approaches ensure that both the semantic intent and factual support of the generated content are traceable and interpretable, thereby supporting semantic consistency across the retrieval and generation pipeline [8]. Modular task decomposition and dynamic agent collaboration further enhance the system's adaptability and robustness, allowing complex information flows and multi-stage decision processes to be coordinated effectively in a retrieval-generation context [9]. Structural priors, modular adapters, and multi-scale feature fusion offer essential methods for compositional integration of external evidence and domain-specific knowledge during both the retrieval and generation stages [10], [11]. These foundations make it possible to maintain core model competencies while flexibly adapting to new data and evolving task demands. Explicit structure-aware decoding mechanisms reinforce this capacity, ensuring that outputs are not only fluent but also strictly anchored to supporting evidence [12].

Long-horizon planning and multi-step reasoning are facilitated by memory-driven hierarchical encoding and dynamic retrieval. These mechanisms empower the framework to track evidence dependencies and sustain coherent reasoning chains, especially in context-rich or sequential generation scenarios [13]. The system's architectural resilience and collaborative reliability are further strengthened through multi-agent orchestration, with specialized modules distributing retrieval, reasoning, and validation to maximize both efficiency and interpretability [14-16].

Advances in neural attention and knowledge augmentation empower explainability and fine-grained evidence control, providing the methodological support required for both accurate attribution and user-facing transparency [17-18]. To ensure practical deployment, parameter-efficient fine-tuning and privacy-preserving adaptation mechanisms enable secure, auditable, and efficient adaptation in sensitive or data-constrained environments [19]. Finally, adaptive policy optimization and risk-aware modeling, drawing on reinforcement learning and uncertainty quantification—complete the methodological toolkit. These foundations are crucial for supporting dynamic adjustment of the retrieval-generation process, improving reliability, verifiability, and alignment with user or application goals [20-21].

In summary, by integrating these core methodological pillars, explicit memory architectures, explainable semantic modeling, modular composition, structure-aware decoding, collaborative multi-agent strategies, and adaptive optimization,this work advances a unified approach to retrieval augmented generation that achieves robust semantic alignment, explicit evidence constraint, and reliable, controllable text generation.

## III. Method

This method generally follows the basic paradigm of retrieval-enhanced generation, but at the modeling level, it simultaneously introduces two core mechanisms: semantic alignment and evidence constraints, to alleviate the structural bias between retrieved information and the generation target. Given an input query, it is first mapped to a continuous semantic representation, which drives subsequent evidence retrieval and generation control. Specifically, the representation of the query in the encoding space is defined as:

$$q = f_{enc}(x) \qquad (1)$$

Here, $f_{enc}(\cdot)$ represents a unified semantic encoding function used to map discrete text to the same semantic space. With this representation, the model no longer relies solely on surface-level term matching but instead uses semantic consistency as a common basis for retrieval and generation, thus providing a unified reference for subsequent alignment and constraints. This paper also presents the overall model architecture, as shown in Figure 1.

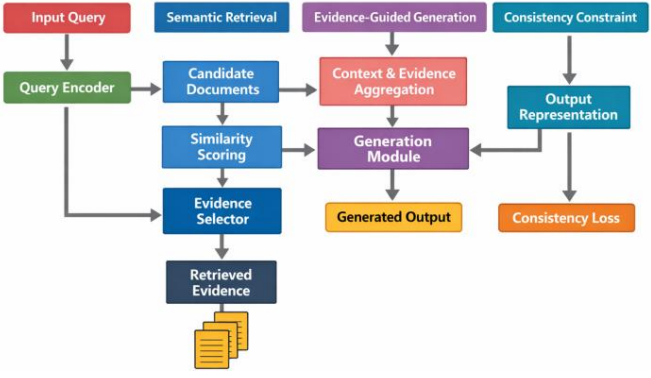

Figure 1. Overall model architecture

During the retrieval phase, explicit semantic alignment modeling is introduced, allowing the relevance between candidate evidence and the query to be characterized by similarity in a continuous space. For any candidate evidence text, its semantic representation can be written as:

$$d_i = f_{enc}(z_i) \quad (2)$$

And based on this, calculate its alignment score with the query:

$$s_i = cos(q, d_i) \quad (3)$$

Here, $cos(\cdot)$ represents the vector cosine similarity. This alignment score measures the semantic consistency between the evidence and the generated target, thereby suppressing retrieval results with large semantic drift from entering the subsequent generation process. Through this mechanism, the retrieval results have undergone preliminary semantic filtering and ranking before entering the generation module.

In the generation phase, the method further introduces an evidence constraint mechanism, transforming the retrieved evidence from implicit context into explicit constraints. When predicting the next symbol at each step, the generative model relies not only on the existing context state but also on the aggregated representation of the evidence semantics.

$$e = \sum_i \alpha_i d_i \quad (4)$$

Where $\alpha_i$ is the weight obtained by normalizing the alignment score. Based on this evidence, the generation probability is modeled as:

$$P(y_t / y_{<t}, e) = Softmax(g(h_t, e)) \quad (5)$$

Where $h_t$ represents the current generation state, and $g(\cdot)$ is the mapping function that integrates the generation state and the evidence information. This modeling method continuously injects evidence information during the generation process, ensuring that the output content is always constrained by the semantic boundaries of the retrieved evidence.

To further strengthen the dependence of generated content on evidence, a consistency constraint is introduced to explicitly suppress deviations between generated semantics and evidence semantics. This constraint can be formalized as:

$$L_{cons} = \|h_{gen} - e\|_2 \quad (6)$$

Here, $L_{cons}$ represents the overall semantic representation of the generated result. By jointly optimizing the generation objective and consistency constraints in the training objective, the model is guided to maintain semantic consistency with the evidence while expressing diversity. Thus, the semantic alignment mechanism is responsible for ensuring the relevance of the selected evidence, while the evidence constraint mechanism ensures that the generation process does not deviate from the factual basis. The two work together within a unified framework to form a controllable and structurally clear retrieval enhancement generation method.

## IV. EXPERIMENTAL ANALYSIS

### A. Dataset

In this study, HotpotQA is selected as the evaluation dataset. The dataset targets knowledge-intensive question answering scenarios. Its core characteristic is that questions usually require information aggregation across multiple evidence fragments to produce complete and verifiable answers. The data are constructed from open encyclopedic corpora. The questions cover diverse information needs, including entities, events, attributes, and relations. This design naturally aligns with the retrieval augmented generation paradigm. It also highlights the necessity of semantic alignment in multi-hop evidence retrieval and the importance of evidence constraints for generation reliability.

Each sample in HotpotQA comprises a question, a ground truth answer, and annotated supporting facts that relate to the answer. These supporting facts are provided at the document, paragraph, or sentence level. The dataset includes predefined training, validation, and test splits, which facilitate reproducible experiments and fair comparisons under consistent settings. Since the dataset explicitly specifies which evidence supports the answer, it offers traceable and verifiable supervision signals. Consequently, it is well-suited for evaluating whether a model genuinely performs evidence-grounded generation rather than relying solely on language priors. Within the scope of this work, HotpotQA can be naturally organized into a query, evidence corpus, and generation triplet. The query corresponds to the question. The evidence corpus consists of encyclopedic documents. The target output is an answer generated in support of the evidence. The semantic alignment module enhances retrieval relevance and multi-hop coherence. It reduces semantic drift and noisy evidence. The evidence constraint module leverages the boundaries defined by the supporting facts. It enforces consistency between the generated content and the retrieved evidence. This enhances answer verifiability and controllability. Such data and task formulation directly align with the design of semantic alignment and evidence constraint mechanisms. It ensures a high degree of consistency between the dataset selection and the research theme.

*B. Experimental Results*

This article first presents the results of the comparative experiments, as shown in Table 1.

Table 1. Comparative experimental results

| Method | EM | F1 | BLEU | ROUGE-L |
|---|---|---|---|---|
| TreeQA[22] | 42.3 | 55.1 | 18.4 | 47.0 |
| CottonBot[23] | 45.7 | 58.9 | 20.1 | 49.3 |
| Vul-rag[24] | 48.2 | 61.4 | 22.7 | 52.6 |
| T-RAG[25] | 51.9 | 65.8 | 25.4 | 55.8 |
| Biorag[26] | 54.6 | 68.2 | 27.1 | 57.4 |
| Ours | 59.8 | 73.5 | 31.6 | 63.2 |

Overall, the proposed method shows consistent advantages across all four metrics, indicating coordinated improvements throughout the retrieval-augmented generation pipeline rather than gains from a single component. Higher EM and F1 reflect more accurate and balanced factual coverage, while simultaneous improvements in BLEU and ROUGE-L show closer alignment with reference wording and structure, suggesting that generation remains within correct evidence boundaries. Compared with baselines, the method continues to improve performance even in saturated settings, demonstrating that its gains stem from tighter integration between retrieval and generation through semantic alignment and evidence constraints, rather than increased capacity alone. These mechanisms reduce noisy evidence, prevent semantic drift, and constrain expression to verifiable content, enabling trustworthy generation that preserves both factual correctness and textual quality; the sensitivity of semantic alignment weights is further analyzed in Figure 2.

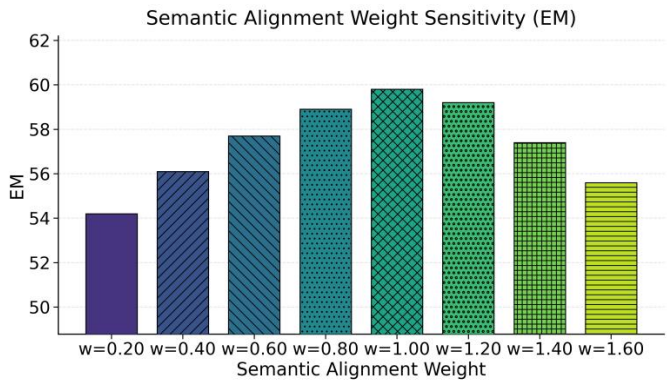

Figure 2. Semantic alignment weights on the EM sensitivity experiment

The results show a non-monotonic relationship between the semantic alignment weight and EM, indicating that this parameter serves as a balancing mechanism rather than a simple performance amplifier. With low weights, weak semantic matching allows loosely related or inconsistent evidence to enter generation, reducing factual accuracy; increasing the weight improves evidence relevance and steadily raises EM. These trends confirm that semantic alignment must balance relevance and coverage to effectively support evidence constraints, acting as a front-end filter that supplies high-quality yet diverse evidence. Moreover, performance remains stable within a reasonable range of weights, highlighting the interpretability and controllability of this parameter; the impact of the Top-K retrieval setting is further analyzed in Figure 3.

An examination of the overall trend shows that the effect of retrieval Top K on EM exhibits clear stage-wise characteristics. This reflects the dual role of candidate evidence size in retrieval augmented generation frameworks. When Top K is small, the amount of evidence passed to the generation module is limited. Information coverage is insufficient. Critical supporting content for answer determination may be missed. This weakens the stability of the generated results under strict matching criteria. This stage highlights the fundamental importance of adequate evidence coverage in knowledge-intensive generation tasks. As Top K gradually increases, the model gains access to a richer set of relevant evidence. The semantic alignment mechanism in the retrieval stage becomes more effective. Highly relevant evidence has a greater chance of being incorporated into the generation. Within this range, the evidence constraint module can perform selection and fusion over a more sufficient candidate set. This allows the generation process to better adhere to verifiable factual expressions. The observed improvement indicates that moderate expansion of candidate evidence helps mitigate bias caused by a single evidence perspective. It leads to more robust generation decisions.

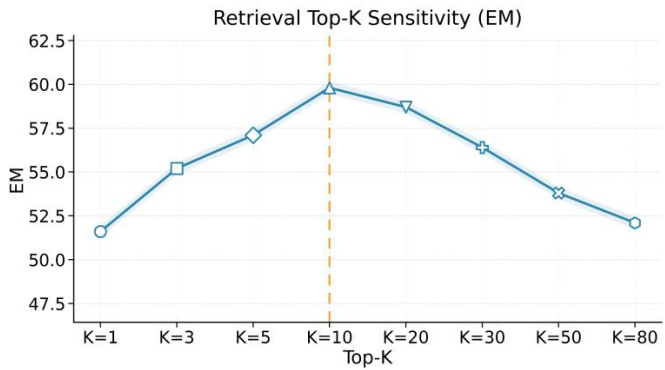

Figure 3. Experiment on the effect of Top-K size on EM sensitivity

When Top K continues to increase, a performance decline is observed. This suggests that an excessive number of candidate evidences significantly raises the noise ratio. The generation stage then faces greater difficulty in evidence selection. Even with semantic alignment and evidence constraint mechanisms, overly redundant evidence sets may introduce fragments that are semantically similar but factually irrelevant. These fragments distract the generation process from key information. This phenomenon demonstrates that a larger retrieval scale is not always better. It must be aligned with the evidence processing capacity of the generation module.

From a methodological design perspective, these sensitivity results further confirm the necessity of fine-grained coordination between retrieval and generation in the proposed framework. A reasonable choice of Top K balances evidence coverage and noise control. Semantic alignment improves candidate evidence quality. Evidence constraints restrict the factual boundaries of generated content. Their joint effect enhances output reliability. At the parameter level, these findings indicate that the proposed method does not rely on extreme settings to be effective. It maintains stability and controllability within a reasonable range. This behavior aligns well with the design goals for high-reliability generation applications.

## V. CONCLUSION

This paper addresses the widely observed issues of insufficient semantic alignment and missing evidence constraints in retrieval augmented generation. It proposes a unified generation framework that integrates semantic consistency modeling with explicit evidence constraint mechanisms. The approach re-examines the relationship between retrieval and generation at a structural level. Retrieved evidence is no longer treated as passive contextual input. Instead, it is explicitly modeled as a core driver of generation decisions. By continuously injecting evidence that is highly relevant to the query and strictly constrained, the model can maintain language fluency while producing content that is more consistently aligned with factual sources. This provides a systematic path toward building trustworthy generation systems. From a methodological perspective, this work emphasizes that improvements in generation quality should not rely solely on stronger language modeling capacity. Greater attention should be paid to whether the generation process is guided by appropriate information boundaries. The semantic alignment mechanism ensures that retrieved evidence remains consistent with generation objectives in semantic space. This establishes a solid foundation for subsequent content integration. The evidence constraint mechanism further limits the generation space. It reduces unsupported inference and information drift. This cooperative design improves the verifiability of generated outputs. At the application level, the proposed framework has significant value for scenarios that require high reliability and traceability. Examples include knowledge-intensive question answering, professional text generation, and decision support tasks. In these settings, generated outputs often require explicit evidence support and coherent logical grounding. By explicitly strengthening the role of evidence constraints during generation, the proposed method helps reduce the risk of incorrect information propagation. It improves system credibility in real-world environments. These properties also provide strong transfer potential. The framework can serve as a general design reference for a wide range of retrieval augmented generation systems.

Looking ahead, retrieval augmented generation methods that combine semantic alignment with evidence constraints are expected to play a greater role in more complex generation scenarios. On the one hand, this framework can be extended to tasks involving multi-evidence aggregation and long-chain reasoning. More refined evidence organization and constraint strategies can further enhance generation reliability for complex problems. On the other hand, as external knowledge sources continue to grow in scale and update frequency, strengthening evidence dependence while preserving generation flexibility will become a key research direction. The framework presented in this paper offers a feasible modeling foundation for this direction. It also provides theoretical and methodological support for building high-trust generation models that align with practical application needs.